%% file: acl_latex.tex
\pdfoutput=1

\documentclass[11pt]{article}

\usepackage[preprint]{acl}

\usepackage{times}
\usepackage{latexsym}

\usepackage[T1]{fontenc}

\usepackage[utf8]{inputenc}

\usepackage{microtype}

\usepackage{inconsolata}
\usepackage{graphicx}
\usepackage{xspace}
\usepackage{amsmath}
\usepackage{amsfonts}
\usepackage{multirow}
\usepackage{booktabs}
\usepackage{colortbl}
\usepackage{tcolorbox}
\usepackage{relsize}
\usepackage{enumitem}
\usepackage{amsmath,amsfonts}

\newcommand{\benchmarkname}[0]{\textsc{GeoHaluBench}\xspace}
\newcommand{\KGname}[0]{\textit{SpatialKG}\xspace}
\newcommand{\algoname}[0]{\texttt{DynamicKTO}\xspace}

\title{Mitigating Geospatial Knowledge Hallucination in Large Language Models: Benchmarking and Dynamic Factuality Aligning}

\author{
\centerline{\bf Shengyuan Wang$^\dagger$, Jie Feng$^\ddagger$, Tianhui Liu$^\S$, Dan Pei$^\P$, Yong Li$^\ddagger$} \\
\centerline{$^\dagger$College of AI, Tsinghua University} \\
\centerline{$^\ddagger$Department of Electronic Engineering, BNRist, Tsinghua University} \\
\centerline{$^\S$School of Electronic and Information Engineering, Beijing Jiaotong University} \\
\centerline{$^\P$Department of Computer Science and Technology, Tsinghua University} \\
\vspace{-0.5cm}
}

\begin{document}
\maketitle
\begin{abstract}
Large language models (LLMs) possess extensive world knowledge, including geospatial knowledge, which has been successfully applied to various geospatial tasks such as mobility prediction and social indicator prediction. However, LLMs often generate inaccurate geospatial knowledge, leading to geospatial hallucinations—incorrect or inconsistent representations of geospatial information—that compromise their reliability. While the phenomenon of general knowledge hallucination in LLMs has been widely studied, the systematic evaluation and mitigation of geospatial hallucinations remain largely unexplored. To address this gap, we propose a comprehensive evaluation framework for geospatial hallucinations, leveraging structured geospatial knowledge graphs for controlled assessment. Through extensive evaluation across 20 advanced LLMs, we uncover the hallucinations in their geospatial knowledge. Building on these insights, we introduce a dynamic factuality aligning method based on Kahneman-Tversky Optimization (KTO) to mitigate geospatial hallucinations in LLMs, leading to a performance improvement of over 29.6\% on the proposed benchmark. Extensive experimental results demonstrate the effectiveness of our benchmark and learning algorithm in enhancing the trustworthiness of LLMs in geospatial knowledge and reasoning tasks. Codes and data are available via \url{https://anonymous.4open.science/r/GeospatialHallucination-823A/}.
\end{abstract}

\input{sec1_intro}
\input{sec2_method}
\input{sec3_experiment}
\input{sec4_related}

\section{Conclusion}
We propose a framework to systematically evaluate and mitigate geospatial hallucinations in LLMs. Using a dedicated taxonomy and controllable evaluation design, we assess 20 advanced LLMs and provide a comprehensive analysis. To improve performance, we enhance the KTO algorithm with dynamic factuality aligning, accounting for geospatial data diversity. With \algoname, smaller open-source models achieve competitive results against top-performing LLMs in hallucination mitigation.

\newpage
\section{Limitations}

\noindent\textbf{To Cover Broader World Knowledge.}
While \benchmarkname currently includes three globally diverse cities—spanning from Asia to America—there are still many regions that should be considered to fully evaluate a world model. Beijing, London, and New York are prosperous cities, but other underdeveloped areas are often less represented, where LLMs may possess less knowledge and exhibit more hallucinations. Additionally, despite \KGname captures key elements of urban space, it can be further enriched with additional information, such as the opening times of POIs, road speed limits, and more. Multi-modal data, including remote sensing or street image services, could also serve as valuable sources of world knowledge. Integrating such data into \KGname and \benchmarkname could provide a more comprehensive understanding and evaluation of the world, beyond just geospatial data.


\noindent\textbf{To Generalize \algoname to More Tasks.}
We have proposed \algoname and demonstrated its superiority in mitigating spatial knowledge hallucination, but further testing is required to benchmark \algoname's performance on other general alignment tasks. Utilizing \algoname to enhance 
LLM's factuality in other domain is also promising. For instance, in the area of law, medicine, finance, etc. there is a strong need for LLMs with less hallucination.

\noindent\textbf{To Explore Various Behaviors.} Teaching an LLM to say "I don't know" is a very exciting and intriguing research question, closely tied to the topic of exploring the knowledge boundaries of LLMs. We explore LLMs behavior of abstaining from answering geospatial knowledge questions, revealing its complexity and potential. Future work should focus on finding a balance between precision and recall when training LLMs to abstain from providing answers.

\section{Ethics Statement}
All the data used in our benchmark and training algorithms come from publicly available sources, including OpenStreetMap\footnote{https://www.openstreetmap.org} and Foursquare\footnote{https://opensource.foursquare.com/os-places/}. We adhere to their respective licenses and have made the code and data available for public access.


\bibliography{custom}

\newpage
\input{sec5_appendix}

\end{document}

%% file: sec1_intro.tex
\section{Introduction}

\begin{figure}
    \centering
    \includegraphics[width=\linewidth]{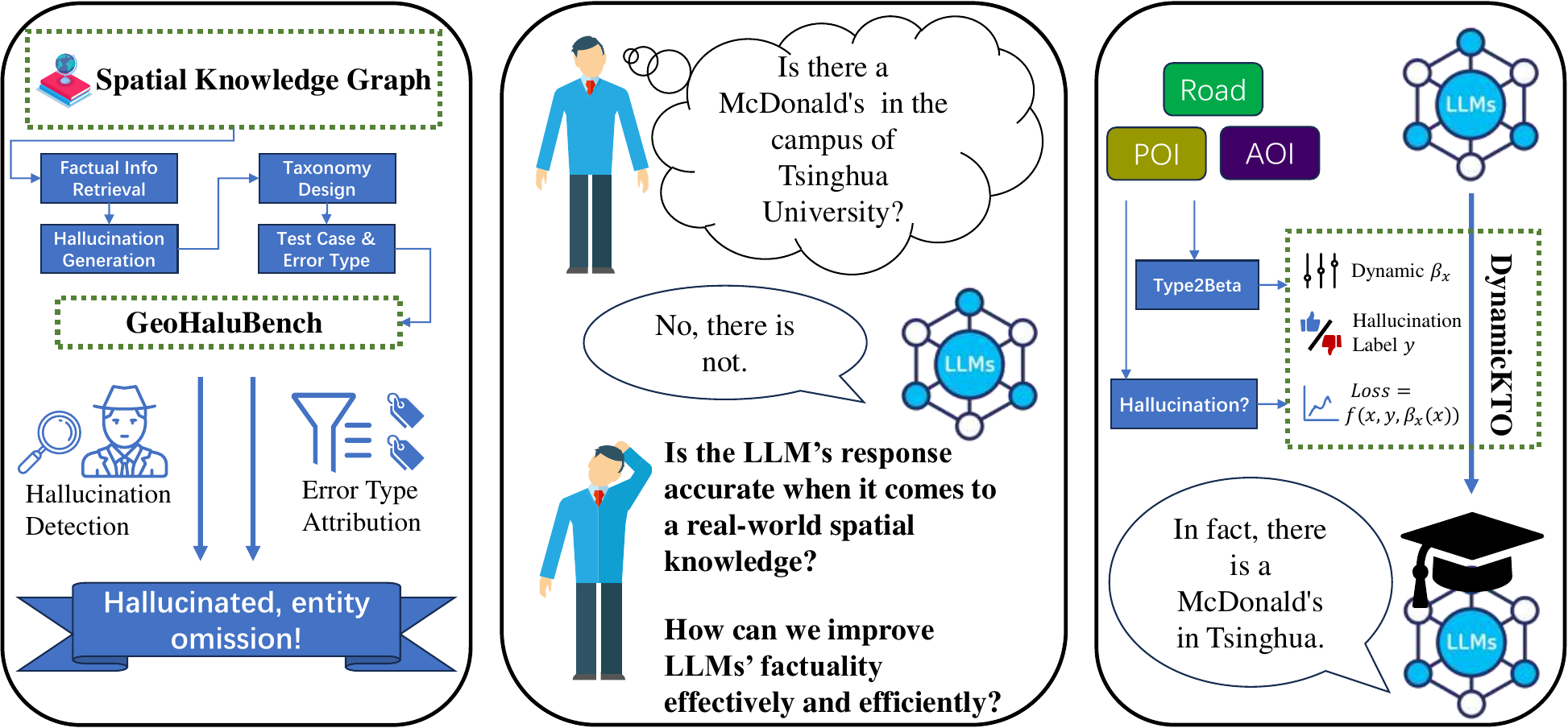}
    \caption{An overview of our work. In order to address real-world geospatial hallucination of LLMs. We propose 1) \benchmarkname to detect and evaluate geospatial knowledge errors, 2) \algoname to enhance LLMs' factuality effectively and efficiently.}
    \vspace{-0.3cm}
    \label{fig:framework}
\end{figure}

Recently, large language models (LLMs), known for their excellent reasoning abilities~\cite{wei2022emergent} and extensive world knowledge~\cite{yu2023kola, ivanova2024elements}, have been widely applied across various domains.
The extensive geospatial knowledge embedded in LLMs has also been explored~\cite{gurnee2023language, roberts2023gpt4geo} and successfully applied to various geospatial tasks. Gurnee et al.\cite{gurnee2023language} and Roberts et al.\cite{roberts2023gpt4geo} demonstrate that LLMs maintain grounded geospatial knowledge that accurately reflects the real world. Leveraging this geospatial knowledge, LLM-based methods have shown promising performance in several geospatial tasks, including GeoLLM for social indicator prediction~\cite{manvi2023geollm}, AgentMove for global mobility prediction~\cite{feng2024agentmove}, and UrbanCLIP for robust and effective urban representation~\cite{yan2024urbanclip}.

While the world knowledge embedded in LLMs has contributed to their widespread success in various applications over the past two years, researchers have identified significant errors and self-contradictions—referred to as hallucinations—in the generated results of LLMs, particularly in domain-specific areas~\cite{huang2023survey, ji2023survey}. These hallucinations significantly affect the trustworthiness of LLMs and their performance in real-world applications. Geospatial knowledge within LLMs also exhibits notable hallucinations in practical use cases~\cite{manvi2024large, feng2024citybench}. Detecting and mitigating these hallucinations has become a critical problem for the development and reliable deployment of LLMs in the real world. 
While various solutions have been proposed for general and domain-specific knowledge hallucinations~\cite{zhang-etal-2024-knowledgeable, chenhalc}, systematic evaluation and mitigation of geospatial knowledge hallucinations remain largely unexplored. This is particularly challenging due to two main factors. First, geospatial data is complex and diverse, resulting in varied manifestations of associated spatial hallucinations. Second, general hallucination reduction methods often fail to account for the unique characteristics of geospatial knowledge, e.g., various elements and relations between them.

In this paper, we propose a systematic benchmark and an effective factuality aligning method to evaluate and mitigate geospatial hallucinations in LLMs. We first build \KGname to reorganize the diverse and unstructured geospatial data, and introduce a set of geospatial evaluation questions, along with a systematic taxonomy of hallucinations. Furthermore, the elements in the evaluation questions are flexible and customizable, allowing for a thorough localization of the shortcomings in geospatial hallucinations across various LLMs. After evaluating 20 advanced LLMs using our proposed benchmark, we find that most of them, especially the open-source LLMs, exhibit significant hallucinations. Based on our observations of these geospatial hallucinations, we develop \algoname to effectively mitigate the geospatial hallucinations in smaller-scale, open-source LLMs by enhancing the KTO algorithm with dynamic factuality aligning. With the help of \algoname, LLama3.1-8B achieves a significant performance improvement on the proposed benchmark and demonstrates competitive performance compared to the second-best model among the 20 advanced LLMs we evaluated. In summary, our contributions are threefold.

\begin{itemize}[leftmargin=1.5em,itemsep=0pt,parsep=0.2em,topsep=0.0em,partopsep=0.0em] 
    \item To the best of our knowledge, we are the first to systematically evaluate and mitigate the geospatial hallucinations in LLMs. 
    \item We have developed a comprehensive benchmark to assess geospatial hallucinations and analyze the performance of 20 advanced LLMs within this framework. 
    \item We propose \algoname, which extends the KTO by incorporating dynamic factuality aligning to account for the diversity and heterogeneity of geospatial knowledge and data. 
    \item Extensive experiments on \benchmarkname and \algoname demonstrate the effectiveness of our proposed framework in evaluating and mitigating geospatial hallucinations. 
\end{itemize}

%% file: sec2_method.tex
\section{Methods}
\subsection{\KGname: Structured Geospatial Knowledge Organization}

As a reliable and informative knowledge base is essential for the definition, detection, and mitigation of the hallucination problem. We construct a high-quality knowledge graph called \KGname based on previous work \cite{liu2023urbankg}. We design a new schema (the high-level structure of KG, including the types of entities and relations) in order to capture fundamental elements in the urban environment and to cover most important relations for geospatial cognition.

In \KGname, fundamental \textbf{entities} include Point of Interest (POI), Area of Interest (AOI) and Road as basic elements describing urban and rural structures. Based on the types of entity in \KGname, we conclude the typical and important \textbf{relations} to describe the spatial connections between entities as follows: POI-LocateAt-AOI, POI-Near-POI, AOI-Near-AOI, AOI-ConnectTo-Road, and Road-Intersect-Road. Mastering the real-world knowledge does not only imply the memory of existing entities' names, but also refers to the capability of recognizing their important attributes. We select the following \textbf{attributes} and link them to the entities in \KGname. For POI, address and category are basic information considered. For regions, attributes include area and type of land use (industrial, residential, etc.). As for Road, we select the length of a road as its attribute discussed.

\KGname is automatically constructed from OpenStreetMap\footnote{https://www.openstreetmap.org} and Foursquare’s Open Source Places\footnote{https://opensource.foursquare.com/os-places/}, which are updating and high-quality city data sources. We design a pipeline to examine and filter the original data for quality control.

\begin{figure}
    \centering
    \includegraphics[width=\linewidth]{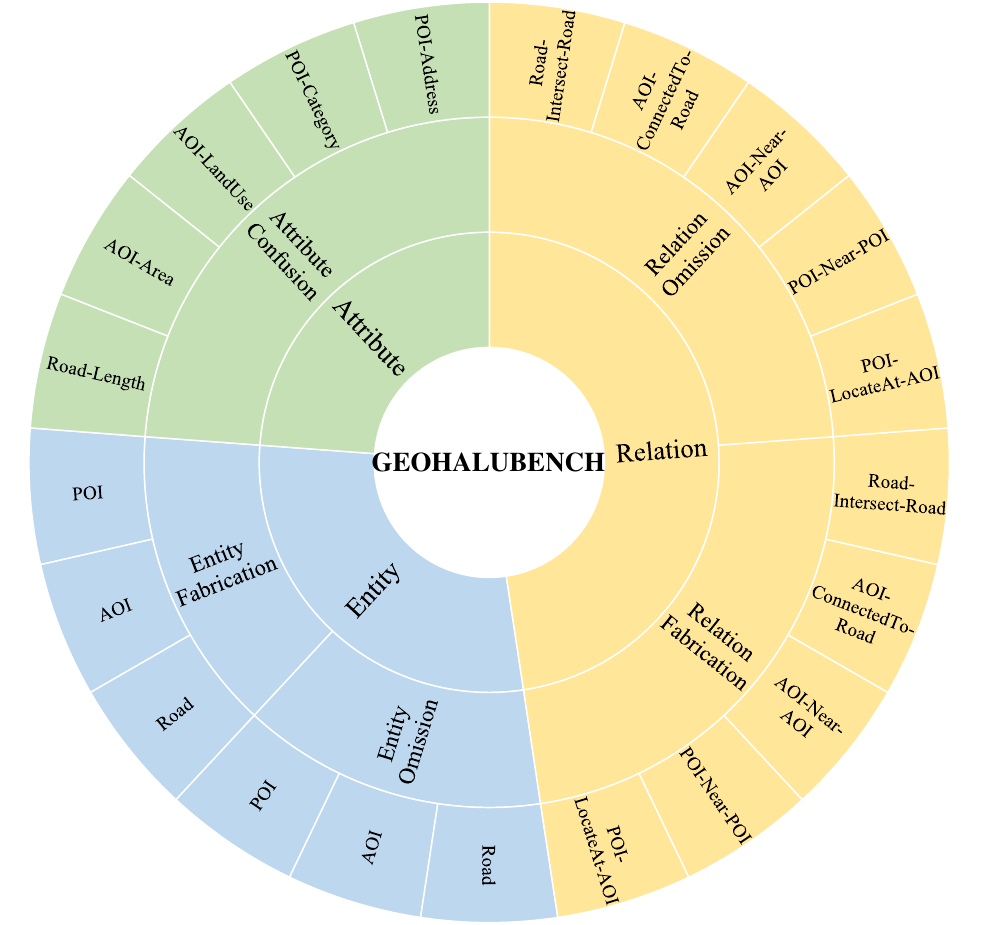}
    \caption{An illustration of the hierarchy and composition of \benchmarkname.}
    \vspace{-0.2cm}
    \label{fig:taxonomy}
\end{figure}

\subsection{\benchmarkname: Geospatial Hallucination Benchmarking}
With adequate knowledge from \KGname, we can classify and benchmark LLM's geospatial  hallucination of real world.
\subsubsection{\benchmarkname Composition}
In this section, we introduce the overall composition of \benchmarkname. As shown in Figure \ref{fig:taxonomy}, we organize data from \KGname into a systematic benchmark with 3 first-level categories and 5 second-level categories. 
\benchmarkname consists of several branches, each corresponding to the regions of a city. Each city includes 21 tasks with a total of 2,100 instances.
\subsubsection{Spatial World Hallucination Taxonomy}
The concept of hallucination traces its roots to the fields of pathology and psychology  \cite{macpherson2013hallucination}. Within the realm of NLP, hallucination is typically referred to as a phenomenon in which the generated content appears nonsensical or unfaithful to the provided source content  \cite{filippova2020controlled, maynez2020faithfulness}. However, existing studies on hallucinations in LLMs typically define hallucination as the generation of incorrect content in terms of factuality or faithfulness \cite{maynez2020faithfulness, ji2023survey, huang2023survey, xu2024hallucination}. 

However, these broad definitions can be vague and insufficient for guiding further and in-depth research for specific fields. To address this gap, we propose a taxonomy for spatial world knowledge hallucination, grounded in knowledge structure. Briefly speaking, geospatial hallucination refers to fabrication of non existing geospatial entities or relations, omission of actual existences, and confusion of their attributions in this work, which is an important subset of hallucination. A more detailed taxonomy within the concept of geospatial hallucination.

\noindent\textbf{Entity-wise.} In this first-level category, we consider the entities in the scenario of hallucination. There are two subcategories that detect different types of hallucinations. \textbf{1) Entity Fabrication: }LLMs will fabricate facts that do not exist actually. \textbf{2) Entity Omission: }LLMs will forget factually existing entities. 

\noindent\textbf{Relation-wise.} Another first-level category roots from the relation among entities. The two subcategories define and examinze the hallucination types in terms of mutual relations between entities. \textbf{1) Relation Fabrication:} This type of hallucination refers to the error of fabricating a factual inaccurate relation between entities. \textbf{2) Relation Omission:} It is a type of hallucination that omits actual relations in the real world.

\noindent\textbf{Attribute-wise.} This category represents a common type of the geospatial hallucination in the real world. \textbf{1) Attribute Confusion.} Even the knowledge about some entities is reliable, there may be errors about its attribute, like category, length, area, etc. in the field of geospatial cognition.

\subsubsection{\benchmarkname Construction}
Every testing sample in \benchmarkname contains a multiple choice question, one reference answer, and a mapping from options to hallucination types. The detailed construction pipeline is described below.

\noindent\textbf{Factual Information Retrieval.} Benchmarking hallucination is based on reliable factual data source. According to the proposed taxonomy of geospatial hallucination, we sample from \KGname in predefined patterns for entity, entity-entity relation, or entity-attribute. They are used as ground truths.

\noindent\textbf{Hallucination Generation.} In each test case, we curate distracting options corresponding to different error types to detect the hallucination type of the testee. These distracting options need hallucination data with given hallucination type. For Entity Fabrication, we first construct non-existing but plausible entities by instructing Meta-Llama-3.1-405B-Instruct. Then, the actual entities in reality are filtered out through comparing with \KGname. For Relation Fabrication category, irrelevant relations within \KGname are created and introduced as hallucinated information. Entity Omission and Relation Omission utilize a void option (None of the others) as the negative option. As for Attribute Confusion, we randomly select values of attribute that is not close to the accurate value than a set threshold.

\begin{figure}[!htbp]
\small
\centering
\caption{An example case of \benchmarkname, where the multiple-choice options are labeled with distinct colors, and the corresponding hallucination types are highlighted in corresponding colors for clarity. 
}
\label{fig:test_case}
\begin{tcolorbox}
\small

\textbf{Test Question} \\
Here is a multiple-choice question: \\
Which of the following is a point of
interest in Beijing? \\
    \textcolor[HTML]{FF5733}{A. Silver Spoon Cafe} \\ 
    \textcolor[HTML]{00CC66}{B. Haidian Library} \\
    \textcolor[HTML]{8A2BE2}{C. None of the other options} \\
Please select from A, B, C. Output your
answer directly 

\tcblower

\textbf{Hallucination Type} \\
\textcolor[HTML]{FF5733}{Hallucinated, Entity Fabrication} \\
\textcolor[HTML]{00CC66}{Factual} \\
\textcolor[HTML]{8A2BE2}{Hallucinated, Entity Omission} 

\end{tcolorbox}
\end{figure}

After factual information retrieval and hallucination generation, the data are transformed into a multiple-choice question, where each option corresponds to a specific type of hallucination or a non-hallucination response. An example is demonstrated in Figure \ref{fig:test_case}.

\subsection{\algoname: Optimization for Hallucination Mitigation}
\subsubsection{Kahneman-Tversky Optimization} 
Kahneman-Tversky Optimization (KTO) \cite{ethayarajh2024kto} is a human-aware loss that directly maximizes the utility of generations inspired by a Kahneman-Tversky model of human utility. Inventors have shown that it matches or exceeds the performance of preference-based methods like Direct Preference Optimization (DPO) \cite{rafailov2023direct} with a more flexible data requirement. 

\subsubsection{\algoname}
In standard KTO, a hyperparameter $\beta \in \mathbb{R}^+$ is introduced to the value function as a control of risk aversion, which serves a similar effect as $\beta$ in the DPO loss, controlling how far $\pi_\theta$ drifts from $\pi_\text{ref}$. However, a fixed beta means the same risk management strategy throughout the dataset, which is not appropriate with varying training data. If the answer to a task is relative easy and fixed, a higher $\beta$ will encourage a closer generation with training samples to avoid risk, resulting in a better performance, vice versa when answers are less certain.
A single, unified $\beta$ value is inadequate for addressing the diverse tasks involved in hallucination mitigation, which is further illustrated by an additional theoretical analysis in Appendix \ref{app:algo_theory}.

Therefore, we propose \algoname, an improved version of the Kahneman-Tversky Optimization (KTO) algorithm for hallucination mitigation, where the hyperparameter $\beta$ is dynamically adjusted. The dynamic $\beta$ is a function of the training sample's feature, allowing more flexible adaptation during the optimization process. The loss function is as follows:
\vspace{-0.3cm}
\begin{equation*}
    L_\text{DynamicKTO}(\pi_\theta, \pi_\text{ref}) = 
     \mathbb{E}_{x,y \sim D} [ \lambda_y - v(x, y) ],
\end{equation*}
\vspace{-0.3cm}
where 
\vspace{-0.1cm}
\begin{equation*}
\small
\begin{split}
    r_\theta(x, y) &= \log \frac{\pi_\theta(y|x)}{\pi_\text{ref}(y|x)}, \\
    z_0 &= \text{KL}(\pi_{\theta}(y'|x)\|\pi_\text{ref}(y'|x)), \\
    \beta (x) &= \text{Type2Beta} (x), \\
    v(x, y) &=
    \begin{cases}
    \lambda_D \sigma(\beta (x)(r_\theta(x,y) - z_0)) \ \text{if } y \sim y_\text{desirable}|x, \\
    \lambda_U \sigma(\beta(x)(z_0 - r_\theta(x,y))) \ \text{if } y \sim y_\text{undesirable}|x.\\
    \end{cases} \\
\end{split}
\end{equation*}
For geospatial hallucination mitigation, $\beta$ is adjusted to 0.1, 0.3, and 0.5 for Entity, Relation, and Attribute respectively.

%% file: sec3_experiment.tex
\section{Experiments}
\input{tables/benchmark_v4}

\subsection{Benchmarking Spatial Hallucination: \benchmarkname}

We systematically evaluate representative LLMs on their geospatial hallucination situation with \benchmarkname.
All evaluations are conducted under zero-shot setting with each model's default prompts. We use the greedy decoding strategy for all LLMs to ensure reproducibility. Following standard practices, Accuracy based on pattern matching is used as the primary metric.

\noindent\textbf{Results: Hallucination Level.} Main results of Beijing are shown in Table \ref{tab:benchmark_v4}. The general performance results are relatively low, revealing significant geospatial hallucinations among different LLMs, even state-of-the-art ones.
Parameter size plays a crucial role in determining the level of hallucination. For instance, within the Qwen family, larger models generally exhibit higher performance when handling spatial world knowledge. However, it is noteworthy that two Qwen2.5 models, with parameter sizes under 3B, perform exceptionally well in this task despite their smaller scale. This suggests that spatial factual tests present a unique challenge compared to other tasks, offering potential for improvement even with more limited model sizes.
The top three LLMs in \benchmarkname are all proprietary models, which require substantial resources for both training and inference. Nevertheless, several open-sourced models have proven competitive, even outperforming GPT-4o.

\begin{figure}[!htbp]
    \centering
    \includegraphics[width=1 \linewidth]{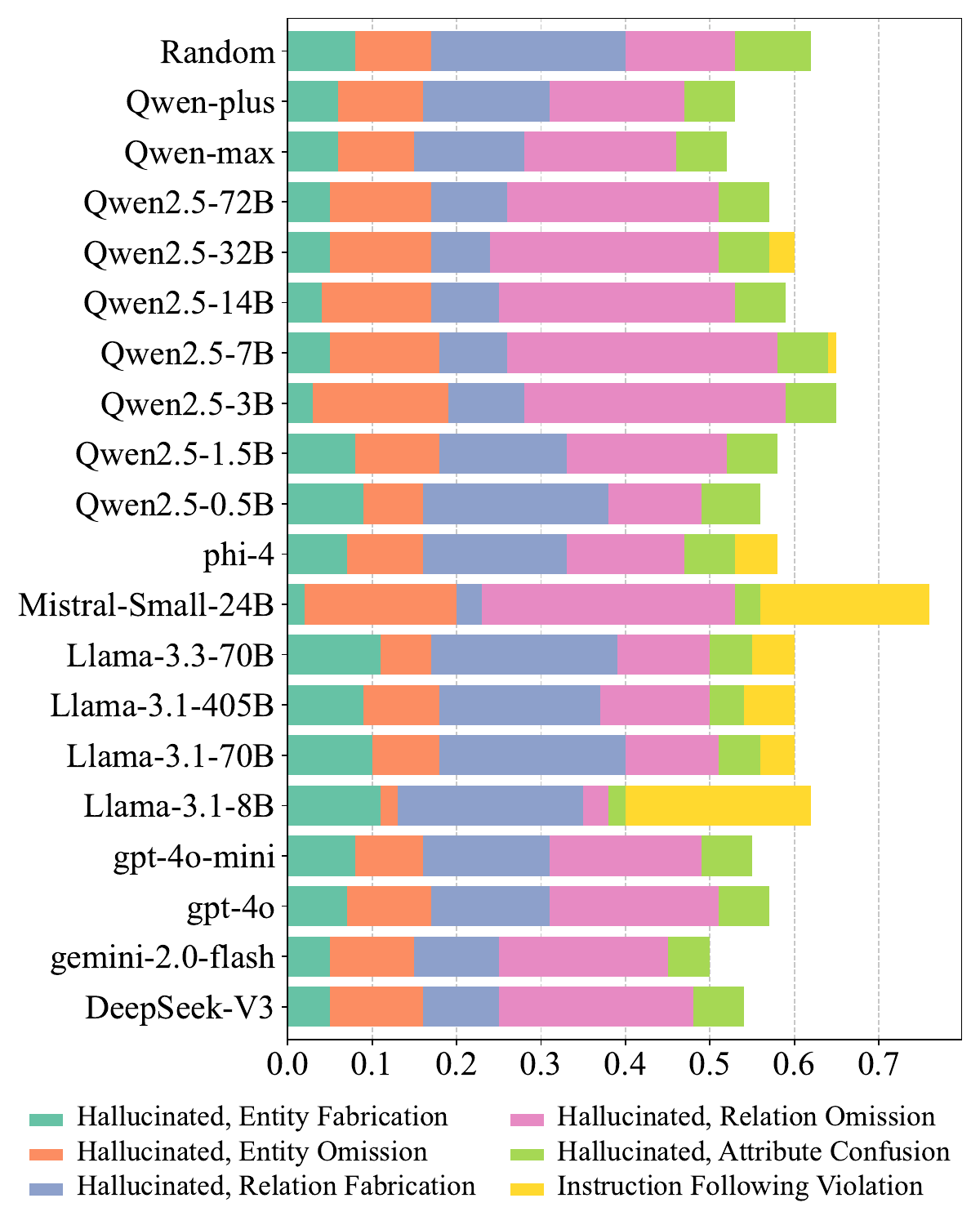}
    \caption{Distributions of Hallucination Types and Instruction Following Violation of different LLMs. Models are listed at name order. All the LLMs are chat models instruction-tuned by their developers.}
    \label{fig:v4_type_distribution}
\end{figure}

\begin{figure*}[!htbp]
    \centering
    \includegraphics[width=\linewidth]{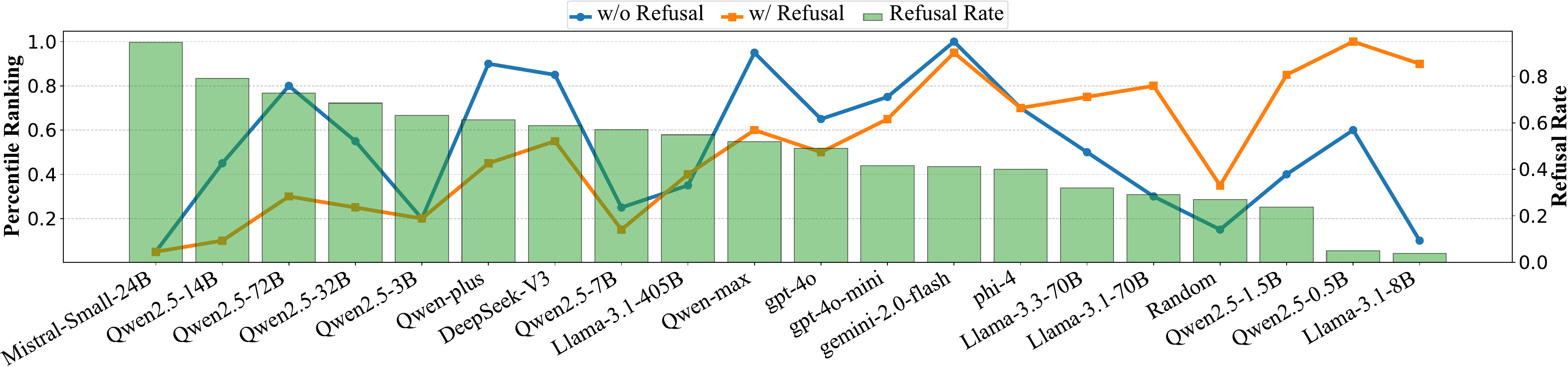}
    \vspace{-0.5cm}
    \caption{The introduction of the Refuse/Abstain option alters the ranking of LLMs. The line chart illustrates the percentile rankings of LLMs on \benchmarkname (without refusal) and \benchmarkname-Abstain (with refusal). Higher points indicate superior rankings and better performance on the respective benchmark.}
    \vspace{-0.2cm}
    \label{fig:ranking_change}
\end{figure*}

\input{tables/benchmark_v4_three_cities}

\noindent\textbf{Results: Hallucination Types.}
The distribution of hallucination types of Beijing is presented in Figure \ref{fig:v4_type_distribution}. The behavior of hallucination varies across different LLMs, depending on factors such as model size and training pipelines. With a few exceptions, most of the LLMs tested are able to follow the instructions in \benchmarkname relatively accurately. 
In general, the ratio of Omission to Fabrication is notably higher than expected, suggesting that the primary cause of geospatial hallucination is a lack of knowledge, rather than the generation of non-factual content. 
In contrast to other model families, the LLaMA series demonstrates a pronounced tendency to fabricate assertions, a tendency that diminishes as the model's parameter size increases.
The Qwen family exhibits a spindle-shaped pattern with respect to Entity/Relation Omission as model size scales, with models ranging from 3B to 32B being more prone to omitting real-world information.
The incidence of hallucinations involving attribute confusion remains relatively high and stable across models, suggesting that this issue either receives limited attention or is inherently difficult to address within current LLM practices.

\noindent\textbf{Results: Multiple Regions.} The results of \benchmarkname for Beijing, London, and New York are presented in Table \ref{tab:three_cities}. Due to space limitations, only representative models and results are included. Across all three cities, the overall performance is relatively low, highlighting the challenges LLMs face in handling global spatial knowledge. Although the absolute accuracy varies between cities, the performance rankings of different LLMs remain highly consistent across all three locations. This consistency demonstrates the robustness and generalizability of the \benchmarkname design. Notably, hallucinations related to New York are systematically less frequent than those for Beijing or London, suggesting a general bias in the world knowledge that current LLMs tend to acquire.

Apart from representative developed cities, we also explore geospatial hallucination situation in underrepresented regions. As shown in Table \ref{tab:three_underrepresented_cities}, the lower performance results demonstrate that LLMs have more tendency to hallucination in less represented areas, showcasing LLMs' less knowledge on these regions.

\noindent\textbf{Extension: Abstain to Answer. }For humans, it is natural to abstain from or refuse to answer a question when faced with a knowledge gap. However, this issue becomes more complex when applied to LLMs. 
While abstaining from answering can help avoid factual errors, it also renders LLMs unreliable or unhelpful as knowledge sources. To evaluate LLMs' hallucination behavior more comprehensively, we incorporate the act of abstaining or refusing to answer into our analysis. 
We expand \benchmarkname to \benchmarkname-Abstain by adding a new option of "Cannot Determine" and conduct an evaluation, with results presented in Figure \ref{fig:ranking_change}. 
Refusing to answer world knowledge questions is a common behavior among LLMs, though the refusal rates vary across different models. This behavior significantly influences performance rankings on \benchmarkname-Abstain, showing a noticeable negative correlation between refusal rate and performance. For instance, LLMs with lower rate of refusal like Llama-3.1-70B, Llama-3.1-8B, Qwen2.5-0.5B or Qwen2.5-1.5B have a huge improvement in ranking. On the other hand, decrease of ranking is generally associated with a high denial rate.
LLMs have different strategies facing the option of abstain originated from different pre-training or post-training process.
When an LLM overuses abstention, it misses opportunities to provide correct answers. Thus, strictly prohibiting or enforcing abstention is neither practical nor reasonable. Effective training should balance precision and recall when teaching LLMs to abstain.

\vspace{-0.2cm}
\subsection{Mitigating Hallucination with \algoname}
\label{sec:Mitigating_Hallucination_with}
Our evaluation on \benchmarkname shows the weakness of current LLMs about the topic of geospatial knowledge in the real world. This situation calls for an effective method to inject LLMs with knowledge about the real world and discourage them from generating hallucinated contents. We implement \algoname and validate its strength compared with existing training methods. Furthermore, we prove that \algoname is not destructive to LLM's original capabilities. Finally, we utilize \algoname with supervised fine-tuning (SFT) to build a more factual model with more accurate world knowledge and capable of various urban spatial tasks.

\noindent\textbf{Datasets and Baselines. }
Due to space limit, please refer Appendix \ref{app:datasets_baselines} for details about datasets and baselines used in experiments.


\input{tables/method_exp}
\noindent\textbf{Results on \benchmarkname.}
Table \ref{method_exp} illustrates the advantage of \algoname in mitigating hallucinations. The model trained with \algoname achieves state-of-the-art (SOTA) performance, or is very close to it, across all three dimensions. When compared to the base model without fine-tuning (Llama3.1-8B-Instruct), a significant reduction in hallucinations is observed. While other fine-tuning or alignment methods generally offer improvements in hallucination mitigation, \algoname outperforms them significantly, as evidenced by the results on \benchmarkname. Furthermore, we have tested various values of $\beta$ in KTO to isolate the effect of hyperparameters. The improvement of \algoname over the best KTO result confirms that the performance boost is not merely due to trivial hyperparameter optimization, but rather stems from the dynamic design of \algoname.

\noindent\textbf{Results on General Benchmarks. }\input{tables/general_bench}Concurrently, \algoname effectively reduces hallucinations while maintaining the model's general capabilities, as evidenced by the results presented in Table \ref{tab:general_bench}. Experiments demonstrate that \algoname is as safe as existing methods. For two of three benchmarks, the degradation of performance is less with \algoname compared with KTO or ORPO. As for MMLU, the drop of performance is around 5\%. In summary, the model maintains its instruction following ability and general knowledge after trained with \algoname.

\input{tables/dynamic_on_two_models}
\noindent\textbf{Model Generalizability.}
As an alignment optimization method, \algoname is model-agnostic and effective across various models in mitigating geospatial hallucinations. Comparison of different models using \algoname is provided in Table \ref{tab:algo_different_model}.

\noindent\textbf{FactualCityGPT. }
\input{tables/fact_citygpt}
Recently, there has been growing interest in enhancing LLMs with real-world cognition and intelligence. Previous work of CityGPT has proposed a LLM with enhanced capabilities on understanding urban space and solving related tasks. However, we observe severe hallucination of spatial knowledge after reproducing and evaluating it, as shown in Table \ref{method_exp}.
To reduce hallucinations, improve its reliability and further enhance LLMs' ability to handle urban tasks, we apply \algoname on CityGPT. As shown in Table \ref{tab:fact_citygpt_results}, the new SFT+\algoname model is still capable of urban spatial tasks, and hallucinate less. Compared to other aliment algorithm baselines, \algoname have significant advantages as well.

%% file: tables/benchmark_v4.tex
\begin{table*}[]

\caption{Results of \benchmarkname on Beijing. For open-source LLMs, the results are presented in descending order of model size. The reported accuracy represents the macro average across the three dimensions in the "Overall" category and the micro average for all other categories. "Ranking" indicates the model's position among the tested LLMs.}
\label{tab:benchmark_v4}
\centering
\resizebox{\linewidth}{!}{

\begin{tabular}{lccccccc}
\toprule
\multicolumn{1}{c}{\textbf{Model}} &
  \textbf{License}  &\textbf{Size}&
  \multicolumn{1}{c}{\textbf{Entity}} &
  \multicolumn{1}{c}{\textbf{Relation}} &
  \multicolumn{1}{c}{\textbf{Attribute}} &
  \multicolumn{1}{c}{\textbf{Overall}} &
  \textbf{Ranking} \\ \midrule
Gemini-2.0-flash &
  Proprietary  &-&
  \textbf{0.4767} &
  \textbf{0.4936} &
  \textbf{0.5400} &
  \textbf{0.5034} & 1\\
Qwen-max-2025-01-25             & Proprietary  &-& 0.4600 & 0.4864 & 0.5160 & 0.4875 & 2\\
Qwen-plus-2025-01-25            & Proprietary  &-& 0.4183 & 0.4840 & 0.5360 & 0.4794 & 3\\
GPT-4o-mini                     & Proprietary  &-& 0.4467 & 0.4520 & 0.4640 & 0.4542 & 6\\
GPT-4o                          & Proprietary  &-& 0.4083 & 0.4344 & 0.4680 & 0.4369 & 8\\
\midrule
DeepSeek-V3                     & Open         &671B& 0.4667 & 0.4528 & 0.4920 & 0.4705 & 4\\
Llama-3.1-405B-Instruct         & Open         &405B& 0.3683 & 0.4216 & 0.4480 & 0.4126 & 14\\
Qwen2.5-72B-Instruct            & Open         &72B& 0.3983 & 0.4304 & 0.5360 & 0.4549 & 5\\
Llama-3.3-70B-Instruct          & Open         &70B& 0.3933 & 0.3952 & 0.4800 & 0.4228 & 11\\
Llama-3.1-70B-Instruct          & Open         &70B& 0.3717 & 0.4016 & 0.4520 & 0.4084 & 15\\
\midrule
Qwen2.5-32B-Instruct            & Open         &32B& 0.4100 & 0.3704 & 0.5040 & 0.4281 & 10\\
Mistral-Small-24B-Instruct-2501 & Open         &24B& 0.2867 & 0.1920 & 0.3280 & 0.2689 & 20\\
Phi-4                           & Open         &14B& 0.4317 & 0.4072 & 0.5040 & 0.4476 & 7\\
Qwen2.5-14B-Instruct            & Open         &14B& 0.3950 & 0.3920 & 0.4800 & 0.4223 & 12\\
\midrule
Llama-3.1-8B-Instruct           & Open         &8B& 0.4183 & 0.3752 & 0.2400 & 0.3445 & 19\\
Qwen2.5-7B-Instruct             & Open         &7B& 0.3550 & 0.3056 & 0.5040 & 0.3882 & 16\\
Qwen2.5-3B-instruct             & Open         &3B& 0.3233 & 0.3272 & 0.4600 & 0.3702 & 17\\
Qwen2.5-1.5B-instruct           & Open         &1.5B& 0.3517 & 0.4328 & 0.4600 & 0.4148 & 13\\
Qwen2.5-0.5B-instruct           & Open         &0.5B& 0.4400 & 0.4552 & 0.3920 & 0.4291 & 9\\
\midrule
\rowcolor[HTML]{EFEFEF} 
Random                          &              -&-& 0.4100 & 0.3880 & 0.2680 & 0.3553 & 18\\
\bottomrule
\end{tabular}
}
\end{table*}

%% file: tables/benchmark_v4_three_cities.tex
\begin{table*}[htbp]
\centering

\caption{Representative results on Beijing, London, and New York. All the LLMs are chat models instruction-tuned by their developers. Results are sorted by model name.}
\vspace{-0.1cm}
\label{tab:three_cities}
\setlength{\tabcolsep}{4pt}
\resizebox{\linewidth}{!}{
\begin{tabular}{lcccc|cccc|cccc}
\toprule
\multicolumn{1}{c}{\multirow{2}{*}{\textbf{Model}}} & \multicolumn{4}{c|}{\textbf{Beijing}}      & \multicolumn{4}{c|}{\textbf{London}}       & \multicolumn{4}{c}{\textbf{New York}}      \\ 
\multicolumn{1}{c}{} &
  \multicolumn{1}{c}{Entity} &
  \multicolumn{1}{c}{Relation} &
  \multicolumn{1}{c}{Attribute} &
  \multicolumn{1}{c|}{Overall} &
  \multicolumn{1}{c}{Entity} &
  \multicolumn{1}{c}{Relation} &
  \multicolumn{1}{c}{Attribute} &
  \multicolumn{1}{c|}{Overall} &
  \multicolumn{1}{c}{Entity} &
  \multicolumn{1}{c}{Relation} &
  \multicolumn{1}{c}{Attribute} &
  \multicolumn{1}{c}{Overall} \\ \midrule
DeepSeek-V3                                & 0.4667 & 0.4528 & 0.4920 & \textbf{0.4705} & 0.4300 & 0.3904 & 0.4080 & 0.4095 & 0.5150 & 0.4336 & 0.4760 & 0.4749 \\
GPT-4o-mini                                & 0.4467 & 0.4520 & 0.4640 & 0.4542 & 0.4767 & 0.4808 & 0.4360 & \textbf{0.4645} & 0.5800 & 0.4784 & 0.4680 & 0.5088 \\
Llama-3.3-70B                     & 0.3933 & 0.3952 & 0.4800 & 0.4228 & 0.4683 & 0.4304 & 0.4720 & 0.4569 & 0.5767 & 0.4472 & 0.5200 & \textbf{0.5146} \\
Llama-3.1-70B                & 0.3717 & 0.4016 & 0.4520 & 0.4084 & 0.4533 & 0.3976 & 0.3920 & 0.4143 & 0.5150 & 0.4416 & 0.4920 & 0.4829 \\
Llama-3.1-8B                      & 0.4183 & 0.3752 & 0.2400 & 0.3445 & 0.3400 & 0.2432 & 0.2640 & 0.2824 & 0.4333 & 0.3360 & 0.2680 & 0.3458 \\
Qwen2.5-72B                       & 0.3983 & 0.4304 & 0.5360 & 0.4549 & 0.4100 & 0.3496 & 0.4400 & 0.3999 & 0.5100 & 0.4256 & 0.5240 & 0.4865 \\
Qwen2.5-7B                        & 0.3550 & 0.3056 & 0.5040 & 0.3882 & 0.3317 & 0.2856 & 0.4080 & 0.3418 & 0.4033 & 0.3384 & 0.5160 & 0.4192 \\ 
\rowcolor[HTML]{EFEFEF} 
Random                                     & 0.4100 & 0.3880 & 0.2680 & 0.3553 & 0.4317 & 0.3704 & 0.3040 & 0.3687 & 0.4100 & 0.3928 & 0.2280 & 0.3436 \\ \bottomrule
\end{tabular}
}
\end{table*}

%% file: tables/method_exp.tex
\begin{table}[!htbp]
\centering
\caption{\algoname outperforms other fine-tuning methods in mitigating hallucinations. The metric used is accuracy, where the accuracy value represents the macro average across three dimensions for the "Overall" category, and a micro average for the remaining categories. Models are evaluated on \benchmarkname-Abstain. The last two rows illustrate the relative improvements.}
\resizebox{1.0\linewidth}{!}{
\begin{tabular}{llcccc}
\toprule
\multicolumn{2}{c}{\textbf{Method}} & \textbf{Entity} & \textbf{Relation} & \textbf{Attribute} & \multicolumn{1}{l}{\textbf{Overall}} \\
\midrule
\multicolumn{2}{l}{Llama3.1-8B} & 0.4050  & 0.3624 & 0.2840 & 0.3505 \\ 
\midrule
\multicolumn{2}{l}{+SFT}           & 0.3833 & 0.3568 & 0.2960 & 0.3454 \\
\multicolumn{2}{l}{+DPO}           & 0.4183 & 0.3768 & 0.2920 & 0.3624 \\
\multirow{3}{*}{+KTO} & $\beta$ = 0.1 & 0.4333 & 0.3912 & 0.3000   & 0.3748 \\
                     & $\beta$  = 0.3 & 0.4300   & 0.3832 & 0.2880 & 0.3671 \\
                     & $\beta$  = 0.5 & 0.4167 & 0.3736 & 0.2920 & 0.3608 \\
\multicolumn{2}{l}{+SimPO}         & 0.4367 & 0.3928 & 0.3120 & 0.3805      \\
\multicolumn{2}{l}{+ORPO}          & 0.4383 & \textbf{0.4320}  & 0.3680 & 0.4128 \\
\midrule
\multicolumn{2}{l}{+\algoname}          & \textbf{0.5717} & 0.4256 & \textbf{0.4600 } & \textbf{0.4858} \\ 
\multicolumn{2}{l}{vs. not fine-tuned} 
& +41.16\% & +17.44\% & +61.97\% & +38.61\% \\
\multicolumn{2}{l}{vs. best KTO} 
& +31.94\% & +8.79\% & +53.33\% & +29.60\% \\
\bottomrule
\end{tabular}}
\vspace{-0.2cm}
\label{method_exp}
\end{table}

%% file: tables/general_bench.tex
\begin{table}[]
\centering
\caption{\algoname does not cause catastrophic interference with the model's general capabilities. We utilize three renowned general benchmarks: IFEval \cite{zhou2023instruction}, BBH \cite{suzgun2022challenging}, and MMLU \cite{hendrycks2020measuring}.
}
\vspace{-0.1cm}
\label{tab:general_bench}
\setlength{\tabcolsep}{4pt}
\resizebox{1\linewidth}{!}{
\begin{tabular}{llcccc}
\toprule

\textbf{Benchmark} & \textbf{Metric} & {\textbf{Llama3.1-8B}} & \textbf{+KTO} & \textbf{+ORPO}   & \textbf{+\algoname} 
\\ 
\midrule
IFEval  & Accuracy & 79.55 & 77.98 & 77.89   & 78.30     \\
BBH   & Score    & 44.33 & 43.82 & 42.84  & 43.43 \\
MMLU    & Accuracy & 69.17 & 68.99 & 68.17 & 63.84                         \\ \bottomrule
\end{tabular}}
\vspace{-0.25cm}
\end{table}

%% file: tables/dynamic_on_two_models.tex
\begin{table*}[]
\centering
\caption{Effects of \algoname with different base models. Both base models are chat models instruction-tuned.}
\vspace{-0.2cm}
\label{tab:algo_different_model}
\resizebox{\linewidth}{!}{
\begin{tabular}{lcccc|cccc|cccc}
\toprule
\multicolumn{1}{c}{\multirow{2}{*}{\textbf{Model}}} &
  \multicolumn{4}{c|}{\textbf{Before Fine-tuning}} &
  \multicolumn{4}{c|}{\textbf{KTO ($\beta=0.1$)}} &
  \multicolumn{4}{c}{\textbf{\algoname}} \\
\multicolumn{1}{c}{}  & Entity & Relation & Attribute & Overall & Entity & Relation & Attribute & Overall & Entity & Relation & Attribute & Overall \\ \midrule
Llama3.1-8B & 0.4050 & 0.3624   & 0.2840    & 0.3505  & 0.4333 & 0.3912   & 0.3000    & 0.3748  & 0.5717 & 0.4256   & 0.4600    & \textbf{0.4858}  \\
Qwen2.5-7B            & 0.2533 & 0.1408   & 0.1480    & 0.1807  & 0.2600 & 0.1400   & 0.1600    & 0.1867  & 0.2967 & 0.2040   & 0.2280    & \textbf{0.2429}  \\ \bottomrule
\end{tabular}
}
\end{table*}

%% file: tables/fact_citygpt.tex
\begin{table*}[htbp]

\caption{We build Factual-CityGPT, an urban spatial LLM with reduced hallucinations. This table presents its performance on hallucination mitigation (\benchmarkname) and urban spatial tasks (CityEval). CI refers to City Image, US to Urban Semantics, SRR to Spatial Reasoning Route, and SRNR to Spatial Reasoning NoRoute.}
\vspace{-0.2cm}
\label{tab:fact_citygpt_results}
\centering
\resizebox{0.9\linewidth}{!}{
\begin{tabular}{l|cccc|cccc}
\toprule
\multicolumn{1}{c|}{\multirow{2}{*}{\textbf{Model}}} & \multicolumn{4}{c|}{\textbf{Hallucination}} & \multicolumn{4}{c}{\textbf{CityEval}} \\
\multicolumn{1}{c|}{} &
  \textbf{Entity} &
  \textbf{Relation} &
  \textbf{Attribute} &
  \textbf{Overall} &
  \textbf{CI} &
  \textbf{US} &
  \textbf{SRR} &
  \textbf{SRNR} \\ \midrule
CityGPT-Llama3.1-8b     & 0.3833    & \textbf{0.3568}   & 0.296   & 0.3454   & 0.5492   & \textbf{0.7000}      & \textbf{0.8440}  & \textbf{0.6460}  \\
Factual-CityGPT-Llama3.1-8b                & \textbf{0.5917}    & 0.3456   & \textbf{0.3760}   & \textbf{0.4378}   & \textbf{0.5569 }  & 0.6933   & 0.8040  & 0.6160  \\ \bottomrule

\end{tabular}
}\vspace{-0.3cm}
\end{table*}

%% file: sec4_related.tex
\vspace{-0.1cm}
\section{Related Work}
\vspace{-0.1cm}
\textbf{Geospatial Knowledge in LLMs}
Trained on large-scale text corpora, LLMs have acquired extensive world knowledge~\cite{ivanova2024elements,yu2023kola}, including global geospatial information~\cite{roberts2023gpt4geo, gurnee2023language}. This embedded geospatial knowledge has inspired the potential application of LLMs in various knowledge-intensive geospatial tasks, such as global geospatial prediction~\cite{manvi2023geollm}—including health, education, and poverty level estimation—mobility prediction~\cite{wang2023would, feng2024agentmove} using text-based addresses, and urban task planning~\cite{jiang2024urbanllm}. However, due to the limitations of online corpora in capturing real-world information, researchers have explored various fine-tuning methods to enhance LLMs' geospatial knowledge, such as CityGPT~\cite{feng2024citygpt} and LAMP~\cite{balsebre2024lamp}. Unlike these studies, which focus on leveraging LLMs' geospatial knowledge for specific tasks, our work is the first to systematically evaluate geospatial knowledge hallucinations and propose an effective mechanism to mitigate them. 

\vspace{-0.1cm}
\noindent\textbf{Hallucination Evaluation of LLMs.}
The hallucination problem in LLMs has been widely studied~\cite{huang2023survey, ji2023survey}, with numerous evaluation benchmarks~\cite{hhem-2.1-open, li2023halueval, liu2023g} and training methods~\cite{ethayarajh2024kto, wubeta} proposed to address general hallucination issues in LLMs. For hallucination evaluation and detection, Niels et al.\cite{mundler2023self} investigate the problem of self-contradiction, while Manakul et al.\cite{manakul2023selfcheckgpt} introduce SelfCheckGPT, a simple sampling-based method to detect hallucinations. Min et al.\cite{min2023factscore} propose FactScore to identify hallucinations in long-form text generated by LLMs. Recently, Ribeiro et al.\cite{ribeiro2022factgraph} and Sansford et al.~\cite{sansford2024grapheval} introduced KG-based frameworks for hallucination detection and evaluation.

\noindent\textbf{Hallucination Reduction of LLMs.}
To mitigate hallucinations in LLMs, one simple approach involves using retrieval-augmented generation (RAG) methods with external knowledge bases during generation~\cite{lewis2020retrieval}. However, RAG-based methods are resource-intensive, requiring a large number of tokens and time during inference. As a result, researchers have continued to explore more efficient methods~\cite{zhang-etal-2024-knowledgeable, tian2023fine, chenhalc} to effectively mitigate hallucinations in various domains. For instance, Zhang~\cite{zhang-etal-2024-knowledgeable} proposes KnowPAT, which constructs a preference set and introduces a new alignment objective for service and urology. Chen et al.\cite{chenhalc} propose HALC, a robust auto-focal grounding mechanism for reducing object hallucinations in vision-language models (VLMs), while Tian et al.\cite{tian2023fine} develop FactTune, a fine-tuning method aimed at reducing hallucinations in biographies and medical queries. 
In contrast to these works focused on general knowledge and domain-specific hallucination evaluation and mitigation, our research specifically targets the evaluation and mitigation of geospatial knowledge hallucinations in LLMs.

%% file: sec5_appendix.tex
\appendix
\section{Appendix}
\label{sec:appendix}

\subsection{Additional Clarification about Geospatial Hallucination's Denotation and Connotation}

We define geospatial hallucination of not only fabrication of non existing entities (relations), but also omission of actual things and confusion of their attributions. From this definition, geospatial hallucination is a subset of hallucination, which can be classified as a subset of 'mistakes' of LLMs.


LLMs' mistake is a broader topic of models generating unsatisfactory contents. There are other kinds of "mistakes" like instruction following violation \cite{zhou2023instruction}, harmful content generation \cite{kumarcertifying}, or misaligned values \cite{wolf2024fundamental}, which are incorrect behaviors of LLMs even with no factual or logical error.

\subsection{Additional Explanation of \algoname's Motivation and Advantages} \label{app:algo_theory}

In this section, we would make extra 
theoretical explanations to our motivation of \algoname from a model's loss perspective.

To the best of our knowledge, we are the first to propse a method of adjusting $\beta$ in KTO and demonstrate its advantages in geospatial hallucination mitigation. 

We conducted analysis with a proposed metric called $\mathcal{L}_{\text{FactDistance}}$ in the following definition, which is inspired by DPO and KTO loss as a way to assess model's judgement between hallucinated and factual information.

Let $\theta$ be the policy (model). Given a pair regarding a same knowledge, their log probalities are defined as follows:

\begin{align}
\label{eq:token_ll}
\log p_{\theta}(\mathbf{x})
  &= \frac{1}{L}\sum_{i=1}^{L}
     \log p_{\theta}\!\bigl(x_i \mid \text{prompt},\,x_{<i}\bigr),\\[4pt]
\log p_{\theta}(\mathbf{y})
  &= \frac{1}{L}\sum_{i=1}^{L}
     \log p_{\theta}\!\bigl(y_i \mid \text{prompt},\,y_{<i}\bigr).
\end{align}

Here, $\mathbf{x}=(x_1,\dots,x_L)$ denotes the factual sequence,  
and $\mathbf{y}=(y_1,\dots,y_L)$ denotes the hallucinated sequence.

\begin{equation}
z^{(n)}
\;=\;
\log p_{\theta}\bigl(\mathbf{x}^{(n)}\bigr)
\;-\;
\log p_{\theta}\bigl(\mathbf{y}^{(n)}\bigr),
\end{equation}

\begin{align}
\sigma(z) &= \frac{1}{1 + e^{-z}}, \\[4pt]
\mathcal{L}_{\text{FactDistance}}
          &= -\frac{1}{N}\sum_{n=1}^{N}\log
             \sigma\!\bigl(z^{(n)}\bigr).
\end{align}

The loss quantifies factual preference by comparing the log probabilities of a factual response against a hallucinated response.

We use the definition above to analyze the influences of training methods to model. The experiment results are shown in Table \ref{tab:fact_distance}. Statistics of $\mathcal{L}_{\text{FactDistance}}$ is demonstrated in Table \ref{tab:fact_distance_stats}.

\input{tables/fact_distance_loss_three_categories}
\input{tables/fact_distance_stats}

The difference of $\mathcal{L}_{\text{FactDistance}}$ among Entity, Relation, and Attribute indicate a difference of difficulty for a LLM to judge and generate factual contents. As a result, a fixed $\beta$ is not appropriate since it enforce same risk management strategy throughout all the data, which calls a dynamic adjusting $\beta$ at task level (Entity, Relation, Attribute). We hope to 1) reduce the overall FactDistanceLoss and 2) enhance stability and consistency.

First, the introduction of the \algoname optimization algorithm leads to a substantial improvement in factual preference, as evidenced by lower FactDistanceLoss scores across all three categories: Entity, Relation, and Attribute.

Second, the standard deviation measures the consistency of the model's performance across different data points. The baseline model has a standard deviation of 0.1472, which decreases with DPO (0.1422) and KTO (0.1386). However, \algoname achieves the most significant reduction to 0.0790, indicating that it not only improves factual preference but also enhances the stability and consistency of the model’s responses across samples.

\subsection{Geospatial Hallucination Results on Underrepresented Regions}

Geospatial hallucination situations of underrepresented regions apart from major cities like Beijing (GDP per capita \$ 30,177), London (GDP per capita \$ 79,069) , or New York (GDP per capita \$119,932) are also worth attention. 

We understand the inherent geographically biases of LLMs introduced by pre-training or post-training stages. The results of geospatial hallucination in large cities is present in Table \ref{tab:three_cities}. Even in most important metropolises, the challenge of hallucination is great. 

We expand our benchmark to include three more underrepresented regions globally, which are Cairo (in Africa, GDP per capita \$8,847), Kabul (in Asia, GDP per capita is \$1,188), and Sucre (in South America, no Wikipedia GDP data, Bolivia's GDP per capita is \$4,014). \footnote{All GDP data above are from the Wikipedia entry of List of cities by GDP or the entry of Economy of Bolivia.} According to various factors, these regions have relative fewer information resources on the Internet, leading to less attention especially in LLMs.

Quantitative evaluation results are shown in Table \ref{tab:three_underrepresented_cities}.
The lower performance results demonstrate that LLMs are facing great challenges of geospatial hallucination especially in underrepresented regions. Compared to results in paper, especially the better performance on New York, we can observe a general bias in the world knowledge that LLMs possess now.

\input{tables/benchmark_underrepresented}

\subsection{Training Cost Analysis of \algoname
}
Through quantitative experiments and complexity analysis, the increase in training cost is manageable. While the introduction of additional operations in \algoname theoretically slows down training, these operations are not computationally intensive. As a result, the impact on training efficiency is acceptable at most of the time. Empirical results show that with 4
 NVIDIA A100 GPUs, standard KTO training takes 3h 1m 49s (181 mins), whereas DynimicKTO training takes 3h 7m 52s (187 mins) —a slight increase of 3.3\%. In addition, since the additional operation is conducted once on each training sample, the training time scales linearly with the amount of data. The approach remains efficient for larger datasets. The slight increase in time cost is almost negligible considering its effectiveness in mitigating hallucinations.

 \subsection{Additional Analysis about \algoname's Performance
}

\algoname's performance on geospatial hallucination mitigation is demonstrate in Table \ref{tab:algo_different_model}. In this section, we would conduct more analysis.

As a novel and competitive alignment method, ORPO is a strong baseline in specific settings. Putting aside the greater data efficiency of KTO, theoretical analysis suggests that if the preference data has sufficiently little noise and sufficiently little intransitivity, then KTO might fall behind of other alignment methods \cite{ethayarajh2024kto}. ORPO is also a advanced algorithm featuring effectiveness, efficiency and scalability \cite{hong2024orpo}. The used factuality alignment dataset, especially the relation subset, is carefully curated, which may bring slight disadvantages for KTO as for relation category (about 1\% accuracy difference). Generally, \algoname is superior than ORPO on the task of geospatial hallucination mitigation. Considering its theoretical potential and empirical superiority, \algoname is especially useful for hallucination mitigation.

\subsection{CityEval Performance Comparision between \algoname and Baselines}
\algoname generalize better to CityEval benchmark than baselines. With the importation of additional finetuning data independent from CityEval tasks, it is expected for factuality aligned models to perform worse than CityGPT. Empirical results demonstrate that \algoname can maintain the most comprehensive capabilities for CityEval while achieving SOTA geospatial hallucination mitigation as demonstrated in Table \ref{tab:fact_citygpt_results_extended}.

\input{tables/fact_citygpt_v2}

 \subsection{Additional Analysis about \algoname's Difference from Sequential Training}
 Previous studies~\cite{lee-etal-2024-instruction,blakeney2024does,leybzon-kervadec-2024-learning} have examined the impact of training data order in language model fine-tuning, highlighting the critical role of data sequencing. In the context of achieving balanced geospatial factual capabilities, replacing DynamicKTO with sequential KTO tuning per task may be suboptimal due to the risk of overfitting to earlier tasks or overwriting previously learned knowledge.

To further investigate this, we conducted experiments using sequential KTO training, applying the same best-performing $\beta$ for each individual task. The results, as shown in Table \ref{naive_sequential}, show that DynamicKTO outperforms naive sequential KTO fine-tuning, reinforcing its advantage in managing knowledge injection across heterogeneous geospatial tasks.

\input{tables/naive_sequential}

\subsection{Automated Taxonomy Initialization of \algoname for Different Tasks}
The core insight behind \algoname originates from the observed heterogeneity in data difficulty across different knowledge categories. However, a fixed $\beta$ means the same risk management strategy throughout the datasets in KTO, which is not appropriate anymore. A better algorithm, adjusting its "tightness" according to training knowledge, is needed for more effective hallucination mitigation.

So, the setting of $\beta$ would be better decided from the statistics of the training data used and domain knowledge. We introduce a metric called $\mathcal{L}_{\text{FactDistance}}$, which quantifies a model's ability to distinguish between factual and hallucinated information. Experimental results in Table \ref{tab:fact_distance} reveal clear differences in judgment difficulty across categories: Entity is the most challenging, followed by Relation, and Attribute is the easiest. Based on this observation, $\beta$ is set inversely proportional to $\mathcal{L}_{\text{FactDistance}}$.

Originally, \algoname is developed specifically to mitigate geospatial hallucination, the primary focus of this study. To that end, we also introduce a Spatial World Hallucination Taxonomy, grounded in both domain understanding and broad geospatial data analysis. The consistency between the taxonomy, the $\mathcal{L}_{\text{FactDistance}}$ analysis, and \algoname’s empirical results further supports the soundness of this task design. 

In terms of tuning granularity, we aim to strike a balance between interpretability, training stability, and data efficiency. Our theoretical and empirical analysis shows that the Entity–Relation–Attribute categorization provides meaningful geospatial distinctions while maintaining training stability.

For domains beyond geospatial knowledge, researchers may want to rely on domain knowledge to set $\beta$ in \algoname. However, within the framework of $\mathcal{L}_{\text{FactDistance}}$
 analysis and \algoname, the $\beta$ can be automatically and easily calculated at different levels, removing the need for predefined task types. Two practical and promising alternatives can be utilized.
 \begin{itemize}
 \item Sample-Level: $\beta$ is calculated individually for each training sample based on its 
 \item Unsupervised Clustering-Level: Clustering with 
 (e.g., via K-Means++) is first applied to the data, and $\beta$ is then computed based on each cluster's average data characteristics.
 \end{itemize}

We also validate these methods for geospatial hallucination. As shown in Table \ref{auto_algo}, experimental results demonstrate significant improvements compared with the baseline method. On the other hand, these two methods still underperform the category-level \algoname. This is mainly due to the instability. So we still recommend using \algoname along with our taxonomy for the geospatial knowledge problem.

\input{tables/automated_algo}

\subsection{Additional Analysis about \benchmarkname's Uniqueness and Advantages over Wikipedia-Based Datasets}
\benchmarkname distinguishes itself among existing Wikipedia-based factual detection datasets. While some shared knowledge may be tested, \benchmarkname has several key distinctions that set it apart.

First, its taxonomy is specially designed for the scenario of geospatial hallucination, whereas other fact-check benchmarks involve general factual accuracy.

Second, unlike wikipedia-based factual datasets like FEVER \cite{thorne-etal-2018-fever} or WiCE \cite{kamoi2023wice}, which contain claims and labels (SUPPORTED, UNSUPPORTED, etc.), \benchmarkname can attribute a test sample’s response to a specific type of hallucination. This feature enhances the interpretability of hallucination detection.

Third, \benchmarkname is automatically constructed from a knowledge graph with data from OpenStreetMap and Foursquare’s Open Source Places, which are high-quality, updating geoinformation services. As a result, GeoHaluBench is competitive in terms of data quality, scalability, coverage, and quantity.

Additionally, with the increasing demand for LLMs to master real-world geospatial knowledge, a speciallized and high-quality benchmark like GeoHaluBench is essential.

Besides, previous related works have examined the possibility of using Wikipedia for geographic problems, but they found limitations of Wikipedia compared to specialized map services like OpenStreetMap. For instance, GeoLM \cite{li-etal-2023-geolm} points out that training with Wikipedia only solves partial challenges in geospatial grounding, as it only provides the linguistic context of a geo-entity with sentences describing history, emographics, climate, etc. The information about the geospatial neighbors of a geoentity is still missing. Another research named GeoLLM \cite{manvi2023geollm} also utilizes map data from OpenStreetMap rather than information from wikipedia.

\subsection{Demonstration of the Dataset}
\subsubsection{Additional Discussion about the Question Form of \benchmarkname}
Multi-choice questions answering serve as the practice to detect hallucination behaviors in \benchmarkname. If a LLM choose a non-existing option like Silver Spoon Cafe in Figure \ref{fig:test_case}, we can reasonable to infer that the LLM 'believes' that Silver Spoon Cafe is an actual POI, demonstrating its vulnerability to hallucination. Besides, multi-choice question can serve as a practical form of interacting with LLMs for its usability, interpretability, and controllability, especially when users regard them as knowledge bases apart from chatbots. We also get inspired by existing hallucination evaluation benchmarks like TruthfulQA \cite{lin2022truthfulqa}, REALTIMEQA \cite{kasai2023realtime}, Med-HAL \cite{pal2023med}, FACTOR \cite{muhlgay2024generating}, etc. which utilize multi-choice QA to detect and analyze various kind of hallucinations, providing valuable assessment for LLMs.

\subsubsection{\benchmarkname Dataset}

Figure \ref{fig:test_case} demonstrates a test case from Entity-POI-Existence category. The original options may contain Chinese characters, as the selected region is Beijing, China; these have been translated into English for demonstration purposes. The situation is similar in other non-English-speaking regions.

\subsubsection{\algoname Fine-tuning Dataset}

\begin{figure}[htbp]
\small
\centering
\caption{A positive and negative training sample used by \algoname.}
\label{fig:train_case}
\begin{tcolorbox}

\begin{center}
\textcolor[HTML]{00CC66}{POSITIVE SAMPLE}
\end{center} 

\textbf{TASK: }[POI\_Category] 

\textbf{USER: }What category does the following POI (Point of Interest) in Beijing belong to: Ajisen Ramen Shuangjing Restaurant?

\textbf{ASSISTANT:} Dining and Drinking > Restaurant > Asian Restaurant > Noodle Restaurant

\textbf{Label:} Factual

\tcblower

\begin{center}
\textcolor[HTML]{8A2BE2}{NEGATIVE SAMPLE} 
\end{center}

\textbf{TASK: }[POI\_Category] 

\textbf{USER: }What category does the following POI (Point of Interest) in Beijing belong to: Ajisen Ramen Shuangjing Restaurant?

\textbf{ASSISTANT: }Dining and Drinking > Restaurant > Fast Food Restaurant

\textbf{Label:} Hallucinated

\end{tcolorbox}
\end{figure}

Figure \ref{fig:train_case} presents two examples used during fine-tuning with \algoname. Although both examples address the same question here, preference pairs are not required. A task label for hyperparameter adjustment, a content label for optimization, and a direction label are sufficient for \algoname.

\subsection{Datasets and Baselines Detail for \algoname Experiments} \label{app:datasets_baselines}

\noindent\textbf{Datasets. }We construct a fine-tuning dataset consisting of 1500 (for entity information) or 2000 (for relation or attribute information) instances of POI, AOI, and Road respectively. These elements are extracted randomly from \KGname and then organized into natural language narratives with templates. From the constructing process, they are annotated to be either hallucinated or factual naturally. In an attempt to generalize, a narrative is paraphrased. To avoid data leakage, none of the instances in the training set is identical to any sample in \benchmarkname.

\noindent\textbf{Baselines. }We compare \algoname with supervised fine-tuning, Direct Preference Optimization (DPO) \cite{rafailov2023direct}, Kahneman-Tversky Optimization (KTO) \cite{ethayarajh2024kto}, Simple Preference Optimization (SimPO) \cite{meng2025simpo}, and Odds Ratio Preference Optimization algorithm (ORPO) \cite{hong2024orpo} for their effects to hallucination mitigation. 
For SFT, we refer the data and training pipeline from CityGPT \cite{feng2024citygpt}, a previous study of injecting LLM with urban knowledge.
We use the standard implementation from LLaMA-Factory \cite{zheng2024llamafactory} for baselines.

\subsection{Prompts and Details of Methods}

We construct hallucinated entities to serve as negative examples for \algoname by instructing Meta-Llama-3.1-405B-Instruct. Figure \ref{prompt_poi}, \ref{prompt_aoi}, \ref{prompt_road} exhibit prompts used in this process.

\begin{figure*}[htbp]
\small
\centering
\caption{The prompt template of generating POI-related hallucinations.}
\label{prompt_poi}
\begin{tcolorbox}
    In a purpose of research, we would like to use imaginary/fictional/mocked information to hallucinate the name of this POI. \\
    Make sure the hallucinated names are natural and realistic as much as possible. They should not be real names. \\
    Please provide five hallucinated names of this POI given the example existing names. \\
    Example existing names: $[real\_poi\_name\_list]$ \\
    Please follow the following format, use [Hallucination] to wrap the hallucinated (generated) names:
    
    [Hallucination] POI Name 1 [Hallucination] 
    
    [Hallucination] POI Name 2 [Hallucination] 
    
    [Hallucination] POI Name 3 [Hallucination] 
    
    [Hallucination] POI Name 4 [Hallucination] 
    
    [Hallucination] POI Name 5 [Hallucination]

\end{tcolorbox}
\end{figure*}

\begin{figure*}[htbp]
\small
\centering
\caption{The prompt template of generating AOI-related hallucinations.}
\label{prompt_aoi}
\begin{tcolorbox}

    In a purpose of research, we would like to use imaginary/fictional/mocked information to hallucinate the name of this AOI.\\
    Make sure the hallucinated names are natural and realistic as much as possible. They should not be real names.\\
    Please provide five hallucinated names of this AOI given the example existing names.\\
    Example existing names: $[real\_aoi\_name\_list]$\\
    Please follow the following format, use [Hallucination] to wrap the hallucinated (generated) names:
    
    [Hallucination] AOI Name 1 [Hallucination]
    
    [Hallucination] AOI Name 2 [Hallucination]
    
    [Hallucination] AOI Name 3 [Hallucination]
    
    [Hallucination] AOI Name 4 [Hallucination]
    
    [Hallucination] AOI Name 5 [Hallucination]

\end{tcolorbox}
\end{figure*}

\begin{figure*}[htbp]
\small
\centering
\caption{The prompt template of generating Road-related hallucinations.}
\label{prompt_road}
\begin{tcolorbox}

    In a purpose of research, we would like to use imaginary/fictional/mocked information to hallucinate the name of this road.\\
    Make sure the hallucinated names are natural and realistic as much as possible. They should not be real names.\\
    Please provide five hallucinated names of this road given the example existing names.\\
    Example existing names: $[real\_road\_name\_list]$\\
    Please follow the following format, use [Hallucination] to wrap the hallucinated (generated) names:
    
    [Hallucination] Road Name 1 [Hallucination]
    
    [Hallucination] Road Name 2 [Hallucination]
    
    [Hallucination] Road Name 3 [Hallucination]
    
    [Hallucination] Road Name 4 [Hallucination]
    
    [Hallucination] Road Name 5 [Hallucination]

\end{tcolorbox}
\end{figure*}

\subsection{Detailed Result Example}

\input{tables/detailed_results}
Table \ref{tab:detailed_results} is the detailed results of \benchmarkname on Beijing at the level of test tasks.

\subsection{Implementation Details}
\subsubsection{Training}
We use LLaMA-Factory \cite{zheng2024llamafactory} for fine-tuning LLMs. As for \algoname, we implement it by modifying LLaMA-Factory. The training epoch is 1 and other key hyperparameters remain same as default except for epoch, batch size, and beta. For experiments in Section \ref{sec:Mitigating_Hallucination_with}, epoch is set to 1. For Factual-CityGPT training, epoch is set to 3.
It takes about 2 hours to train a 8B model for 1 epoch with 8 $\times$ A100 GPUs.
\subsubsection{Evaluation}
Opencompass\footnote{0.3.9 version} \cite{2023opencompass} is used for our evaluation on general benchmarks, all tested models are deployed locally with lmdeploy \cite{2023lmdeploy}.

For \benchmarkname, we deploy our fine-tuned models and LLaMA-3.1-8B with VLLM \cite{kwon2023efficient}. The temperature is set to 0 for reproducibility. Other parameters are as default. The rest LLMs are used via APIs.

\subsection{Case Study}

We use the instance in Figure \ref{fig:test_case} as an example. According to reliable knowledge sources, Haidian Library is an existing point of interest (POI) in Haidian District, Beijing. In contrast, Silver Spoon Cafe is not a POI but a fabricated name. If the LLM selects option A, "Silver Spoon Cafe," it mistakenly believes the cafe is located in Beijing, which exemplifies Entity Fabrication Hallucination. On the other hand, if the LLM selects option C, "None of the other options," it incorrectly rules out the other two options as valid entities, thereby overlooking the real POI. This represents Entity Omission Hallucination.

%% file: tables/fact_distance_loss_three_categories.tex
\begin{table}[!htbp]
\centering

\caption{$\mathcal{L}_\text{FactDistance}$ changes after training. Base model is the chat model instruction-tuned by its developer.}
\label{tab:fact_distance}
\resizebox{1.0\linewidth}{!}{
\begin{tabular}{lcccc}
\toprule
\textbf{Category} & \textbf{Llama3.1-8B} & \textbf{+DPO} & \textbf{+KTO} & \textbf{+DynamicKTO} \\
\midrule
Entity    & 1.1072 & 1.0874 & 1.0770 & 0.7209 \\
Relation  & 0.9465 & 0.9372 & 0.9278 & 0.8200 \\
Attribute & 0.7471 & 0.7401 & 0.7383 & 0.6265 \\

\bottomrule
\end{tabular}
}
\end{table}

%% file: tables/fact_distance_stats.tex
\begin{table}[!htbp]
\centering

\caption{Training with \algoname\ lowers both the mean and the variability (standard deviation) of $\mathcal{L}_{\text{FactDistance}}$.}
\label{tab:fact_distance_stats}
\resizebox{1.0\linewidth}{!}{
\begin{tabular}{lcccc}
\toprule
\textbf{Metric} & \textbf{Llama3.1-8B} & \textbf{+DPO} & \textbf{+KTO} & \textbf{+DynamicKTO} \\
\midrule
Macro Average    & 0.9336 & 0.9216 & 0.9144 & 0.7225 \\
Standard Deviation  & 0.1472 & 0.1422 & 0.1386 & 0.0790 \\

\bottomrule
\end{tabular}
}
\end{table}

%% file: tables/benchmark_underrepresented.tex
\begin{table*}[htbp]
\centering

\caption{Representative results on Cairo, Kabul, and Sucre. All the LLMs are chat models instruction-tuned by their developers. Results are sorted by model name.}
\vspace{-0.1cm}
\label{tab:three_underrepresented_cities}
\setlength{\tabcolsep}{4pt}
\resizebox{\linewidth}{!}{
\begin{tabular}{lcccc|cccc|cccc}
\toprule
\multicolumn{1}{c}{\multirow{2}{*}{\textbf{Model}}} & \multicolumn{4}{c|}{\textbf{Cairo}}      & \multicolumn{4}{c|}{\textbf{Kabul}}       & \multicolumn{4}{c}{\textbf{Sucre}}      \\ 
\multicolumn{1}{c}{} &
  \multicolumn{1}{c}{Entity} &
  \multicolumn{1}{c}{Relation} &
  \multicolumn{1}{c}{Attribute} &
  \multicolumn{1}{c|}{Overall} &
  \multicolumn{1}{c}{Entity} &
  \multicolumn{1}{c}{Relation} &
  \multicolumn{1}{c}{Attribute} &
  \multicolumn{1}{c|}{Overall} &
  \multicolumn{1}{c}{Entity} &
  \multicolumn{1}{c}{Relation} &
  \multicolumn{1}{c}{Attribute} &
  \multicolumn{1}{c}{Overall} \\ \midrule
DeepSeek-V3                                & 0.4883 & 0.3304 & 0.4440 & 0.4209 & 0.4817 & 0.3360 & 0.4280 & 0.4152 & 0.4900 & 0.3328 & 0.5000 & 0.4409 \\
GPT-4o-mini                                & 0.4917 & 0.3904 & 0.4040 & 0.4287 & 0.5600 & 0.3960 & 0.4400 & 0.4653 & 0.5533 & 0.4048 & 0.4760 & 0.4780 \\
Llama-3.3-70B                     & 0.4900 & 0.3840 & 0.4640 & 0.4460 & 0.5983 & 0.3632 & 0.4360 & 0.4658 & 0.5250 & 0.3672 & 0.4720 & 0.4547 \\
Llama-3.1-70B                     & 
0.4717 & 0.3568 & 0.4080 & 0.4122 & 0.5583 & 0.3680 & 0.4240 & 0.4501 & 0.5200 & 0.3656 & 0.4400 & 0.4419 \\
Llama-3.1-8B                      & 0.4150 & 0.3696 & 0.2400 & 0.3415 & 0.3500 & 0.2984 & 0.2240 & 0.2908 & 0.4000 & 0.3704 & 0.2480 & 0.3395 \\
Qwen2.5-72B                       & 0.4367 & 0.3384 & 0.4640 & 0.4130 & 0.4067 & 0.3216 & 0.4600 & 0.3961 & 0.4617 & 0.3336 & 0.4960 & 0.4304 \\
Qwen2.5-7B                        & 0.3900 & 0.2778 & 0.4160 & 0.3613 & 0.3450 & 0.2989 & 0.4120 & 0.3520 & 0.4183 & 0.2834 & 0.4160 & 0.3726 \\ 
\rowcolor[HTML]{EFEFEF} 
Random                                     & 0.4367 & 0.4024 & 0.2680 & 0.3690 & 0.4167 & 0.3867 & 0.2040 & 0.3358 & 0.3933 & 0.3752 & 0.2720 & 0.3468 \\ \bottomrule
\end{tabular}
}
\end{table*}

%% file: tables/fact_citygpt_v2.tex
\begin{table}[htbp]

\caption{\algoname can generalize better to CityEval benchmark than other baselines. CI refers to City Image, US to Urban Semantics, SRR to Spatial Reasoning Route, and SRNR to Spatial Reasoning NoRoute. +\algoname refers to Factual-CityGPT-Llama3.1-8B, which is finetuned on CityGPT-Llama3.1-8B with \algoname.}
\vspace{-0.2cm}
\label{tab:fact_citygpt_results_extended}
\centering
\resizebox{\linewidth}{!}{
\begin{tabular}{lcccc}

\toprule

\multicolumn{1}{c|}{\multirow{2}{*}{\textbf{Model}}} & \multicolumn{4}{c}{\textbf{CityEval}} \\
\multicolumn{1}{c|}{} &
  \textbf{CI} &
  \textbf{US} &
  \textbf{SRR} &
  \textbf{SRNR} \\ \midrule
CityGPT-Llama3.1-8B     & 0.5492   & \textbf{0.7000}      & \textbf{0.8440}  & \textbf{0.6460}  \\
+KTO     & 0.4077   & 0.5233      & 0.7880  & 0.6280  \\
+DPO     & 0.4338   & 0.5567      & 0.7840  & 0.6200  \\
+\algoname                & \textbf{0.5569 }  & 0.6933   & 0.8040  & 0.6160  \\ \bottomrule

\end{tabular}
}\vspace{-0.3cm}
\end{table}

%% file: tables/naive_sequential.tex
\begin{table}[!htbp]
\centering
\caption{\algoname cannot replaced by naively model training with sequential KTO with different hyperparameters. ST denotes Sequentially Trained. MT denotes Mixed Trained.}
\resizebox{1.0\linewidth}{!}{
\begin{tabular}{llcccc}
\toprule
\multicolumn{2}{c}{\textbf{Method}} & \textbf{Entity} & \textbf{Relation} & \textbf{Attribute} & \multicolumn{1}{l}{\textbf{Overall}} \\
\midrule

\multicolumn{2}{l}{ST w/ KTO}           & 	0.4567 & 0.4168 & 0.3960 & 0.4232 \\
\multicolumn{2}{l}{MT w/ KTO}           & 0.4333 & 0.3912 & 0.3000 & 	0.3748 \\

\multicolumn{2}{l}{\algoname}         & \textbf{0.5717} & \textbf{0.4256} & \textbf{0.4600} & \textbf{0.4858}      \\

\bottomrule
\end{tabular}}
\label{naive_sequential}
\end{table}

%% file: tables/automated_algo.tex
\begin{table}[!htbp]
\centering
\caption{The predefined taxonomy is helpful for geospatial hallucination mitigation. The number of clusters is set to three for fair comparison. S. Level denotes Sample-Level. UC. Level denotes Unsupervised Clustering-Level. C. Level denotes Category-Level.}
\resizebox{1.0\linewidth}{!}{
\begin{tabular}{llcccc}
\toprule
\multicolumn{2}{c}{\textbf{Method}} & \textbf{Entity} & \textbf{Relation} & \textbf{Attribute} & \multicolumn{1}{l}{\textbf{Overall}} \\
\midrule

\multicolumn{2}{l}{+KTO}           & 0.4333 & 0.3912 & 0.3000 & 	0.3748 \\

\multicolumn{2}{l}{+\algoname (S. Level)}           & 	0.4400 & 0.4336 & 0.4160 & 0.4299 \\

\multicolumn{2}{l}{+\algoname (UC. Level)}         & 0.4600 & \textbf{0.4730} & \textbf{0.4080} & \textbf{0.4467}      \\

\multicolumn{2}{l}{+\algoname (C. Level)}         & \textbf{0.5717} & 0.4256 & \textbf{0.4600} & \textbf{0.4858}      \\

\bottomrule
\end{tabular}}
\label{auto_algo}
\end{table}

%% file: tables/detailed_results.tex
\begin{table*}[]
\centering
\caption{Detailed results of \benchmarkname on Beijing. PE refers to POI-Existence, AE refers to AOI-Existence, RE refers to Road-Existence, PLoA refers to POI-LocateAt-AOI, PNeP refers to POI-Near-POI, ANeA refers to AOI-Near-AOI, ACoR refers to AOI-ConnectTo-Road, RCoR refers to Road-ConnectTo-Road, PAddr refers to POI-Address, PCate refers to POI-Category, ALand refers to AOI-LandUse, AArea refers to AOI-Area, RLeng refers to Road-Length. Models are in original names in APIs.}
\resizebox{\linewidth}{!}{
\begin{tabular}{llllllllllllll}
\hline
\multicolumn{1}{c}{\textbf{Model}} &
  \multicolumn{1}{c}{\textbf{PE}} &
  \multicolumn{1}{c}{\textbf{AE}} &
  \multicolumn{1}{c}{\textbf{RE}} &
  \multicolumn{1}{c}{\textbf{PLoA}} &
  \multicolumn{1}{c}{\textbf{PNeP}} &
  \multicolumn{1}{c}{\textbf{ANeA}} &
  \multicolumn{1}{c}{\textbf{ACoR}} &
  \multicolumn{1}{c}{\textbf{RCoR}} &
  \multicolumn{1}{c}{\textbf{PAddr}} &
  \multicolumn{1}{c}{\textbf{PCate}} &
  \multicolumn{1}{c}{\textbf{ALand}} &
  \multicolumn{1}{c}{\textbf{AArea}} &
  \multicolumn{1}{c}{\textbf{RLeng}} \\ \hline
DeepSeek-V3                     & 0.180 & 0.255 & 0.600 & 0.332 & 0.084 & 0.024 & 0.260 & 0.368 & 0.200 & 0.800 & 0.520 & 0.040 & 0.100 \\
gemini-2.0-flash                & 0.170 & 0.335 & 0.640 & 0.360 & 0.076 & 0.196 & 0.388 & 0.516 & 0.160 & 0.820 & 0.600 & 0.240 & 0.380 \\
gpt-4o                          & 0.135 & 0.290 & 0.505 & 0.320 & 0.104 & 0.116 & 0.268 & 0.308 & 0.160 & 0.680 & 0.460 & 0.080 & 0.200 \\
gpt-4o-mini                     & 0.270 & 0.390 & 0.485 & 0.360 & 0.124 & 0.144 & 0.324 & 0.264 & 0.140 & 0.760 & 0.540 & 0.040 & 0.200 \\
Llama-3.1-8B-Instruct           & 0.385 & 0.375 & 0.455 & 0.452 & 0.320 & 0.240 & 0.408 & 0.392 & 0.040 & 0.640 & 0.500 & 0.200 & 0.040 \\
Llama-3.1-8B-Instruct           & 0.385 & 0.375 & 0.455 & 0.452 & 0.320 & 0.240 & 0.408 & 0.392 & 0.040 & 0.640 & 0.500 & 0.200 & 0.040 \\
Llama-3.3-70B-Instruct          & 0.320 & 0.345 & 0.475 & 0.476 & 0.196 & 0.108 & 0.300 & 0.276 & 0.180 & 0.840 & 0.580 & 0.020 & 0.120 \\
Meta-Llama-3.1-405B-Instruct    & 0.160 & 0.265 & 0.510 & 0.324 & 0.096 & 0.140 & 0.252 & 0.160 & 0.100 & 0.840 & 0.600 & 0.000 & 0.000 \\
Meta-Llama-3.1-70B-Instruct     & 0.290 & 0.330 & 0.460 & 0.468 & 0.232 & 0.276 & 0.328 & 0.204 & 0.280 & 0.800 & 0.640 & 0.000 & 0.040 \\
Mistral-Small-24B-Instruct-2501 & 0.030 & 0.075 & 0.095 & 0.008 & 0.000 & 0.000 & 0.000 & 0.000 & 0.000 & 0.360 & 0.120 & 0.000 & 0.000 \\
phi-4                           & 0.225 & 0.325 & 0.475 & 0.352 & 0.068 & 0.128 & 0.368 & 0.320 & 0.200 & 0.800 & 0.620 & 0.080 & 0.200 \\
qwen2.5-0.5b-instruct           & 0.395 & 0.465 & 0.475 & 0.484 & 0.444 & 0.372 & 0.496 & 0.480 & 0.160 & 0.600 & 0.480 & 0.320 & 0.260 \\
qwen2.5-1.5b-instruct           & 0.235 & 0.240 & 0.515 & 0.440 & 0.184 & 0.156 & 0.348 & 0.352 & 0.320 & 0.760 & 0.480 & 0.180 & 0.300 \\
Qwen2.5-14B-Instruct            & 0.055 & 0.105 & 0.365 & 0.132 & 0.004 & 0.000 & 0.088 & 0.032 & 0.120 & 0.700 & 0.220 & 0.000 & 0.000 \\
Qwen2.5-32B-Instruct            & 0.085 & 0.165 & 0.465 & 0.164 & 0.020 & 0.008 & 0.136 & 0.156 & 0.200 & 0.760 & 0.420 & 0.000 & 0.040 \\
qwen2.5-3b-instruct             & 0.165 & 0.300 & 0.450 & 0.292 & 0.032 & 0.008 & 0.040 & 0.064 & 0.020 & 0.740 & 0.180 & 0.000 & 0.000 \\
Qwen2.5-72B-Instruct            & 0.085 & 0.180 & 0.480 & 0.184 & 0.024 & 0.028 & 0.140 & 0.192 & 0.120 & 0.780 & 0.420 & 0.020 & 0.020 \\
Qwen2.5-7B-Instruct             & 0.125 & 0.160 & 0.405 & 0.236 & 0.020 & 0.008 & 0.092 & 0.064 & 0.100 & 0.700 & 0.260 & 0.100 & 0.160 \\
qwen-max-2025-01-25             & 0.165 & 0.300 & 0.575 & 0.292 & 0.068 & 0.152 & 0.412 & 0.336 & 0.200 & 0.860 & 0.480 & 0.020 & 0.040 \\
qwen-plus-2025-01-25            & 0.115 & 0.245 & 0.555 & 0.240 & 0.024 & 0.128 & 0.216 & 0.304 & 0.220 & 0.800 & 0.560 & 0.100 & 0.100 \\
Random                          & 0.290 & 0.285 & 0.315 & 0.220 & 0.324 & 0.244 & 0.296 & 0.304 & 0.240 & 0.260 & 0.240 & 0.200 & 0.220 \\ \hline
\end{tabular}
}
\label{tab:detailed_results}
\end{table*}